\documentclass[sigconf,natbib=true]{acmart}

\usepackage{tikz}
\usepackage{bbm}
\usepackage{natbib}
\usepackage[show]{notes-alt}
\usepackage{soul}
\usepackage[utf8]{inputenc}
\AtBeginDocument{%
  \providecommand\BibTeX{{%
    \normalfont B\kern-0.5em{\scshape i\kern-0.25em b}\kern-0.8em\TeX}}}
    
\newcommand{\expred}{ExPred}
\newcommand{\hotflip}{HotFlip}
\newcommand{\polyjuice}{Polyjuice}
\newcommand{\counterassist}{SparCAssist}
\newcommand{\mlm}{Masked Language Model}

\copyrightyear{2022}
\acmYear{2022}
\setcopyright{acmcopyright}\acmConference[SIGIR '22]{Proceedings of the 45th
International ACM SIGIR Conference on Research and Development in Information
Retrieval}{July 11--15, 2022}{Madrid, Spain}
\acmBooktitle{Proceedings of the 45th International ACM SIGIR Conference on
Research and Development in Information Retrieval (SIGIR '22), July 11--15, 2022,
Madrid, Spain}
\acmPrice{15.00}
\acmDOI{10.1145/3477495.3531677}
\acmISBN{978-1-4503-8732-3/22/07}

\begin{document}
\fancyhead{}

\title{SparCAssist: A Model Risk Assessment Assistant Based on Sparse Generated Counterfactuals}






\author{Zijian Zhang}
\email{zzhang@l3s.de}
\orcid{0000-0001-9000-4678}
\affiliation{%
  \institution{L3S Research Center}
  \streetaddress{Appelstr. 9a}
  \city{Hannover}
  \state{Lower Saxony}
  \country{Germany}
  \postcode{30167}
}
\author{Vinay Setty}
\orcid{0000-0002-9777-6758}
\email{vsetty@acm.org}
\affiliation{%
  \institution{L3S Research Center}
\country{Germany}
}
\affiliation{%
    \institution{University of Stavanger}
    \city{Stavanger}
    \country{Norway}
 }

\author{Avishek Anand}
\email{avishek.anand@tudelft.nl}
\orcid{0000-0002-0163-0739}
\affiliation{
  \institution{L3S Research Center}
\country{Germany}
}
\affiliation{%
  \institution{Delft University of technology}
  \city{Delft}
  \country{Netherlands}
}

\begin{abstract}
We introduce \counterassist{}, a general-purpose risk assessment tool for the machine learning models trained for language tasks.
It evaluates models' risk by inspecting their behavior on counterfactuals, namely out-of-distribution instances generated based on the given data instance.
The counterfactuals are generated by replacing tokens in rational subsequences identified by \expred, while the replacements are retrieved using \hotflip~or \mlm~based algorithms.
The main purpose of our system is to help the human annotators to assess the model's risk on deployment.
The counterfactual instances generated during the assessment are the by-product and can be used to train more robust NLP models in the future.
\end{abstract}



\maketitle
\section{Introduction}
\label{sec:intro}

In this paper we present a demonstration of \counterassist{}: a tool for human-assisted debugging of machine learning (ML) models for language tasks.
Debugging machine learning models in terms of estimating if a trained model is fit for deployment is considered more of an art than a science with a lot depending on the heuristics employed by the model developer ~\cite{bhatt2020ExplainableMachineLearninginDeployment}.
Standard deployment checks often involve heavy reliance on performance metrics on validation and test sets to evaluate the generalization and robustness of systems. 
That means there is heavy focus on inspecting incorrectly predicted instances.
We argue that ensuring good performance is not enough.
Deep neural networks for language tasks are known to be particularly fragile on out-of-distribution instances even if they deliver impressive validation set accuracy~\cite{ross2020explaining}.

Some of the known risks of deploying such over-parameterized models such as transformers is that they tend to memorize undesirable shortcuts in the training data that lead to reasonable validation performance but leads to failures in real-world scenarios~\cite{zheng2021irrationality}.
Unlike common practice of ML metrics, we propose to provide a system that uses counterfactuals as examples to help the user, in this case the model developer, to understand the risks of the model under consideration.
Specifically, given a trained ML model, our intention is to allow the user to identify instances that have potential risk during deployment. 
An instance is considered~\textit{risky} if its minor human-understandable modifications result in undesirable predictions.  

Consider a \textit{document classification task} of classifying review text of movies into \texttt{positive} and \texttt{negative} sentiments. 
Say a trained sentiment classifier classifies the input ``the movie was awesome'' as a positive sentiment.
However, by slightly perturbing it to a counterfactual sentence ``the movie \st{was} \textit{is} awesome'' results in a negative sentiment. 
This counterfactual sentence exposes two potential risks in the model, firstly, the model focuses on non-essential words towards making a prediction and secondly, the model is not robust under non-decisive perturbation.
Other approaches that provide insights into the reasons behind predictions of complex machine learning models, a large variety of post-hoc explanation methods have been proposed~\citep{lime,lundberg2017unified,integratedgradients}.
However, all these explanation approaches operate on existing instances on the training or the test set.
Therefore they cannot effectively assess the potential risk when the models are deployed in the field.

The aim of \counterassist{}~is to modify existing counterfactual generation approaches to help humans identify instances that might be particularly risky. 
We allow users enough control to choose input instances that are of interest and pieces of text out of them within that might be important for the task under consideration.  
In allowing a fine-grained control to the inputs choosing and feedback providing, the users are able to both get a sense of the functioning of the model as well as be educated about the potential risks it might possess. 
In~\counterassist{}, we work on the task of sentiment classification as an example but in principle \counterassist{}~can be extended to a large set of language tasks, for example question answering, fact checking, textual entailment etc.

The system then provides multiple ``\textit{perturbations}'' of the input sentence that result in a change in the prediction.
The perturbations should be small enough so that we can effectively assess the model's local risk, specifically in the close vicinity surrounding a given instance.
Therefore, we are only interested in \textit{sparse} perturbations or we only allow a small number of perturbations to the original input. 
The users' role is to assess whether the generated counterfactual is \textit{plausible, natural} and if the model's prediction is \textit{correct} on counterfactuals.
Plausible here means that the input text is believed to be written or conceived by a human.
In terms of approach we start from a trained model and identify sentences that need to be scrutinized.
We choose this subset of sentences using \expred\cite{expred}, a select-then-predict system that is interpretable by design.
We then use adversarial approaches like~\hotflip{}~\cite{ebrahimi-etal-2018-hotflip} and \mlm{} approaches to generate counterfactuals. 
These generated counterfactuals are then shown to humans who in turn interact with the suggestions to establish its plausibility and validity.
Finally, we compute the \textit{risk} of the model under consideration from the trail of user interactions.

The primary goal is to assess the model's deficiencies using the perturbations, and then our secondary goal is to generate a dataset of counterfactual explanations, which are grammatically plausible and meaningful from human perspective, could be the by-product of the system.


\subsection{Related Work}
\label{sec:relatedworks}

We divide the related work into two parts -- counterfactual generation, and automated approaches for model debugging. 
The counterfactual generation approaches can be categorized into two classes: automatic approaches and human-annotation approaches.
Automatic approaches for generating counterfactuals are predominantly based on gradient-based optimization to generate adversarial examples like~\hotflip{}~\cite{ebrahimi-etal-2018-hotflip}, AutoPrompt~\cite{shin2020autoprompt}, and Imitation Attacks \cite{wallace-etal-2020-imitation}. 
Another set of approaches to automatically generate counterfactuals is to use prompt-based approaches like~\polyjuice{}~\cite{wu2021polyjuice}.
However, the main disadvantage of automatic approaches is that the generated counterfactuals are not natural, i.e., likely to be not  plausible from human perspective.
Prompt-based approaches are slightly more natural that gradient-based approaches, but they rely on heuristic prompts that have to be manually provided.

On the other hand, some approaches conduct user-study with crowd workers to obtain human-understandable counterfactual datasets ~\cite{mcdonnell2016relevant, kaushik2019learning}.
These approaches are understandably non-scalable and exhibit high variance due to subjectivity and the well-known Rashomon effect~\cite{davis2015rashomon}.
We use both gradient-based and prompt-based approaches for counterfactual generation but instead use humans-in-the-loop methods for quality assessment.
More precisely, we take a hybrid approach for producing counterfactuals with utility of the counterfactual as the most desirable outcome. 

The second piece of related work is regarding debugging ML models that have been used in text classification~\cite{monarch2021human}, search~\cite{singh2018posthoc,singhKWA21:valid:l2r,singh2020model}, and many language tasks in general~\cite{idahl2021towards,funke2021zorro} by using explanations or interpretable machine learning approaches called explanation-based human debugging (EBHD).
\citet{10.1162/tacl_a_00440} recently review EBHD approaches that exploit explanations to enable humans to give feedback and debug NLP models.
Unlike adversarial algorithms~\cite{misra2019black, wallace-etal-2020-imitation} the goals of these systems is to explicitly expose model's deployment risk by ensuring that models are \emph{right for the right reason}.
In addition to academic efforts, model debugging tools like~\cite{zhang:faxplain,Singhexs2019:exs},  SageMaker\footnote{\url{https://docs.aws.amazon.com/sagemaker/latest/dg/train-debugger.html}} provided by Amazon are also available.
Other human-in-the loop systems include~\cite{singh2016discovering,holzmann2016tempas} to update models and knowledge bases.
However, they mostly focus on training process and do not provide a post-hoc model risk assessment.

\section{Solution Approach}
\label{sec:approach}
The primary goal of our \counterassist~tool is to assist users to assess the deployment risk of the NLP classification model trained on specific datasets using counterfactual instances (counterfactuals).
The input to \counterassist~is a model together with  a dataset using which the risk assessment of the model is to be performed.
The \counterassist~generates counterfactuals based on a selected data instance and tries to flip the model's prediction for that instance.
The tool assesses the risks of the model based on the user's feedback on the plausibility and meaningfulness of the counterfactuals.
The secondary output \counterassist~tool is plausible and meaningful counterfactual dataset for future training.
In this section, we state our problem by first defining the counterfactual formally with its three essential properties and then formalize the model's risk.

\subsection{Interpretation Using Counterfactuals}
Interpretation-by-Counterfactuals, also known as interpretation using minimal contrastives~\cite{ross2020explaining} is a model-interpretation technique that allows humans to understand the reason behind prediction by investigating models' behavior under counterfactual inputs~\cite{MILLER20191, lipton_1990}.
According to~\cite{ross2020explaining,hilton2017social}, in order to assess the risk of the models, the counterfactuals need to satisfy following rules:
\begin{itemize}
    \item[1] The number of replaced tokens should be as few as possible.
    \item[2] The counterfactual instance should be plausible and meaningful~\cite{chomsky2009syntactic}.
\end{itemize}
Like in \hotflip~\cite{ebrahimi-etal-2018-hotflip} and \polyjuice~\cite{wu2021polyjuice}, we mainly focus on the counterfactual generated by replacing the tokens.
These two approaches, however, do not constrain the number of replaceable tokens while generating the counterfactuals.
This will result in an overwhelming number of alternative counterfactuals for human annotators.
In \counterassist, we address this issue by firstly focusing on sentences containing rationale subsequences identified by the \expred~model~\cite{expred}.
We then only allow \emph{rationale} tokens identified by \expred~in the selected sentence to be replaced, while the other tokens are kept fixed.
Alternatively to the \expred, users can manually select the tokens allowed to be replaced in the rationale sentences.

More formally, we use  $\mathbf{x} = [x_0, x_1, \cdots, x_l]$ to represent the input sequence (rationale sentence) to our tool and the model to be assessed is represented as $\mathcal{M}$, while it's predicted label for a given instance $x$ is denoted as $\hat{y} = \mathcal{M}(\mathbf{x} \odot \mathbf{m})$, where $\mathbf{m}\in \{0, 1\}^l$ is the rationale mask, $\odot$ stands for element-wise multiplication. We then define a counterfactual generation function $C_{x_i\to v}$ which replaces the  rationale token $x_i$ with a counterfactual token $v$ from the 
vocabulary $V$. A counterfactual with only one token replaced is called \textit{1-counterfactual} and represented as $\mathbf{c}^1_{\mathbf{x}} = C_{x_i\to v| i\in \mathbb{1}(m), v\in V}(\mathbf{m} \odot \mathbf{x})$.
The condition $i\in\mathbb{1}(m)$ restricts the tokens to be replaced to the rational mask $\mathbf{m}$.
We repeat the above processes until the prediction on the counterfactual sentence changes.


\subsection{Model Risk Assessment}
We consider a model as ``risky'', when its behavior on the counterfactuals doe not match domain experts' (users') decision, despite its high accuracy on the given, or inside-distribution, instances, including instances from the train, dev-test, and test set~\cite{koyama2020invariance}. 
Since the global risk is usually intractable~\cite{ribeiro2018anchors}, we consider its local counterpart instead, namely how wrong does the model predict on the counterfactuals sampled from the vicinity of an instance in the dataset.
\begin{figure}
    \centering
    \includegraphics[width=5cm]{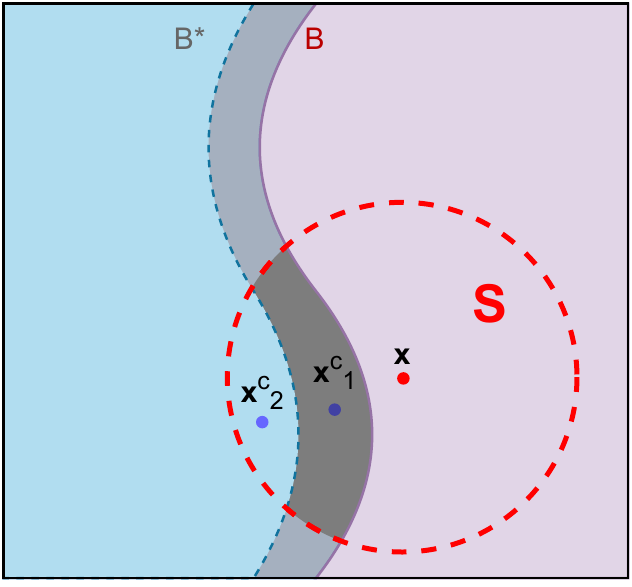}
    \caption{Risk and unexpected behavior in the vicinity of an inside-distribution-instance $\mathbf{x}$. The dark gray area is the risky area.}
    \label{fig:risk}
\end{figure}

As illustrated in Fig. \ref{fig:risk}, for a counterfactual $\mathbf{x}^c$ sampled from the vicinity of an inside-distribution instance $\mathbf{x}^c\in S(\mathbf{x})$, if the decision of the model $\mathcal{M}(\mathbf{x}^c)$ is dissonant with the decision of domain experts', we call this an ``risky behavior'' and the area resulting in risky behavior ``risky volume''.
The local risk score of the model is the ratio between risky volume and the volume of $S$.
In more detail, we assume an ``optimal'' decision boundary $B^*$ as a hyperplane resulting in decisions aligning with domain experts' prediction, while $B$ is the decision boundary of the model.
The model's local risk in $S$ is then the ratio between the volume of $V$ and $\mathcal{S}$, where $V$ is the interior enclosed by $S$ and both decision boundaries as $V$, namely $\partial{V}=(B^*\cup B)\cap S$.
We can estimate it by dividing the number of samples resulting in risky behaviors with the number of samples drawn universally from $S$ using the Monte-Carlo strategy.
However, it is difficult to draw samples from such vicinity due to the discreetness of human language sentences.
Therefore, instead of drawing counterfactual samples universally, we focus on such counterfactuals resulting in the model's prediction flipping.
Such examples locate right across the model's decision boundary and will most likely expose the model's risk.

For example, the sentiment classifier should predict $\mathbf{x}=$``I love this movie.'' as positive and the counterfactuals $\mathbf{x}^c_2=$``I \st{love}~\emph{hate} this movie.'' to negative.
Because this is in accord with users' decisions, the model's risk does not increase.
But if the model predicts $\mathbf{x}^c_1=$``I like this \st{movie}~\emph{film}.'' to negative, the risk score increases, as users will not agree with this prediction.
We ask the annotator to rate the prediction ($f\in{1-5}$, $f$ for faithfulness) by asking to how much extent does the model predicts in concord with her expectation.
We do not ask for a binary rating because the optimal decision boundaries are vague and subjective.
Then for each annotator, we calculate her risk score around the instance $\mathbf{x}$ by: $\text{risk} = \sum_{\mathbf{x}^c\in\mathcal{X}^c}(5-f(\mathbf{x}^c))/|\mathcal{X}^c|$, where $\mathcal{X}^c$ represents the set of counterfactual samples provided by this annotator.
The final risk of the model in $S$ is estimated by the weighted average of risk scores from all annotators, where the weight is the number of the counterfactuals generated in her/his session.
It is worth mentioning that it requires further research to decide the exact number of counterfactuals needed to effectively evaluate the model's risk.

\vspace{-0.1cm}
\section{System Design}

\begin{figure*}
    \centering
    \includegraphics[width=\linewidth]{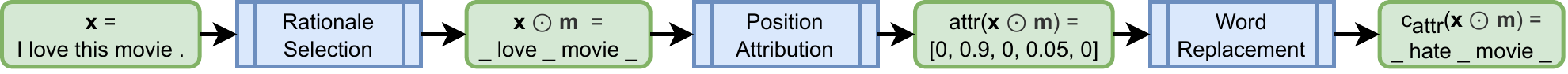}
    \caption{The system diagram of the counterfactual generation assistant}
    \label{fig:system}
\end{figure*}
\begin{figure*}
    \centering
    \includegraphics[width=\textwidth]{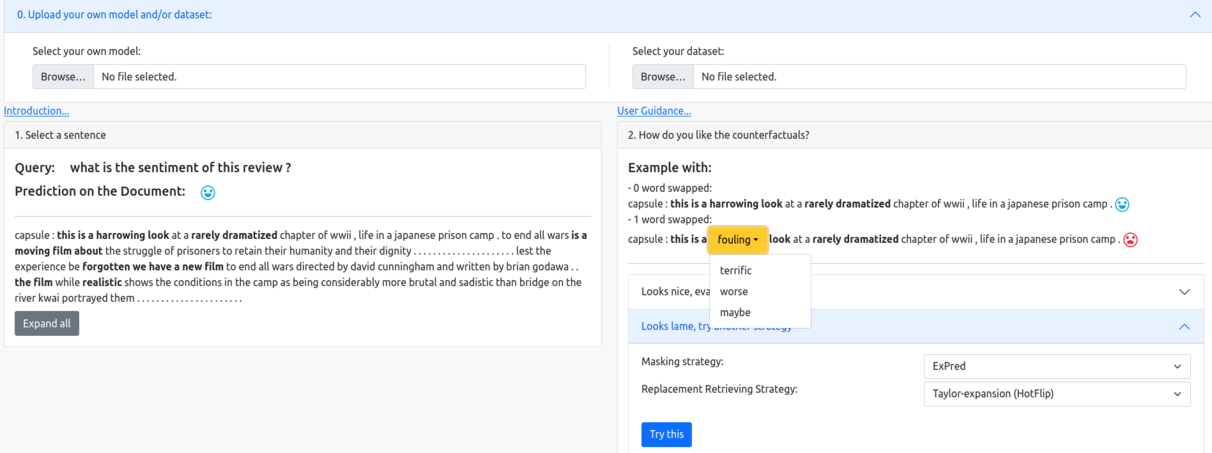}
    \caption{The screenshot of the \counterassist~user interface}
    \label{fig:screenshot}
\end{figure*}
As illustrated in Figure \ref{fig:system}, we decompose the \counterassist~ system into three components: \textbf{Rationale Selection}, \textbf{Position Attribution} and \textbf{Word Replacement}. 
The \counterassist~generates counterfactuals are as follows:

\begin{enumerate}

\item  It first randomly selects an instance $\mathbf{x}$ from the given dataset which may or may not be labeled. For the demo, we use the ``Movies'' datasets for the sentiment analysis task~\cite{deyoung-etal-2020-eraser}.

\item  It then leverages the \expred~model~\cite{expred} which is a two-phase select-then-predict model pre-trained on the movies dataset to select the \textbf{rationale tokens}.
\item  For all those rationale tokens, we \textbf{attribute their positions} and select candidate position(s) based on the attribution.
In our \counterassist, we are only interested in such counterfactuals changes the model's prediction ($\mathcal{M}(\mathbf{c}) \neq \hat{y}$), therefore:
\item  It \textbf{replaces the tokens} on each selected position so that the estimated loss regarding the alternative label other than the original prediction is minimized.
\item  It then selects the final single counterfactual with only one token replaced based on their actual prediction loss.
\item  We repeat steps 3 to 5 until the prediction changes or the number of word replacements exceeds a pre-defined upper limit.
Based on our empirical observation, the set the upper limit to five replacements.
\end{enumerate}

We integrate two counterfactual generation approaches to \counterassist, the Taylor-expansion based \hotflip~and a \mlm~(MLM)-based algorithm used in \polyjuice.
These two approaches are decomposed into two phases, namely \textbf{position attribution} and \textbf{token selection} as step 3 and 4 introduced above. We explain these two approaches below:

\paragraph{\textbf{\hotflip~approach:}} For this approach, we follow~\cite{ebrahimi-etal-2018-hotflip} to search for positions and words using the beam search. More formally. we attribute positions with the dot-product of corresponding token embedding with the partial derivative of the loss regarding the alternated prediction label with respect to the token embedding, i.e. $\frac{\partial}{\partial x_i}l(\hat{y}, \bar{\hat{y}} \neq \hat{y})\cdot \mathbf{e_i}$, where $\mathbf{e_i}$ represents the embedding of the i-th token.
    We select the top-$p$, $p\leq r$ of the position, where $r$ is the number of rationale tokens.
    Then in token selection, for each select position we choose the token that maximizes the dot-product of the new token's embedding with the partial derivation, i.e. $v_{replace} = argmax_{v\in V}\frac{\partial}{\partial x_i}l(\hat{y}, \bar{\hat{y}} \neq \hat{y})\mathbf{e^v}$.
    
\paragraph{\textbf{Masked Language Modelling-based approach:}} For this method, we first construct the MLM model's input with the prompt template as suggested in ~\cite{ptuning_liu,autoprompt,prompt_tuning_lseter} as follows: \emph{``<masked sequence>'' the sentiment of this review is ``<alternative label>''}. For example, \emph{``this movie is \texttt{[mask]}'' the sentiment of this review is ``negative''}.
    
    Unlike in the hotflip, where the positions are attributed with the product of gradient and the embedding, there is no such close-form position attribution in the MLM-based method.
    Therefore we trivially attribute all tokens with the same score
    For the token replacement, we mask each rationale word and apply the prompt template, then predict the masked word using a pre-trained blank-filling model based on RoBERTa~\cite{liu2019roberta} (with pre-trained weights from the Huggingface framework\footnote{\url{https://huggingface.co/roberta-large}}). For each word position, we keep the token with the highest score, then we select the final counterfactual which minimizes the loss with respect to the alternative label.

A screenshot of our system is illustrated in Fig. \ref{fig:screenshot}.
The top part shows widgets to upload the model and the corresponding dataset. The lower part is split into two columns:

1. The left part (\textbf{``1. Select a sentence''}) presents a document selected from the dataset, where rationale tokens are highlighted with boldface font.
By default, all sentences containing at least one rationale token are displayed.
The users can expand the full review by clicking the ``Expand all'' button.
Even after restricting the sentences using the rationale masks, the length of the whole review document is still large, users can therefore select a single sentence at a time to inspect its counterfactuals.

2. In the right hand side, (\textbf{``2. How do you like the counterfactuals?''}) the counterfactuals are displayed following the order of word replacement, with a replaced word highlighted.
As introduced in the previous section, the users are asked to evaluate the \textit{plausibility} and \textit{meaningfulness}, together with the \textit{faithfulness} of the prediction, if the user is satisfied with the current counterfactual generation.
The users may also choose to use the default \expred~rationale masks or provide a custom mask by selecting the alternative rationale tokens according to their preference.
\vspace{-0.1cm}
\section{Conclusion and Future Work}
In this paper, we propose \counterassist, an ML model debugging tool. \counterassist~allows users to construct word-replacement-based counterfactual instances of rationale masked sentences. These counterfactuals are helpful for better understanding of model's behavior and identifying deployment risks of the underlying the model.
The \counterassist~prunes the choices of counterfactuals by restricting the range of variable tokens to rationale tokens only. 
This minimizes the number of counterfactual choices presented to the users.
Optionally, the users can also specify custom masks for the variable tokens.
We implement \hotflip~and \mlm-based approaches to retrieve out-of-distribution (OOD) tokens and enable the users to evaluate generation results.
It is also possible to deploy the \counterassist~system and gather counterfactual annotations with evaluations from large crowd-workers through web browsers.
Another advantage of our tool is that it is flexible and generic. It allows debugging any ML model with a user chosen dataset. 
It easily supports replacing the rationale-masking, position attribution and OOD tokens selection components with other methods. 

As for the future work, as mentioned in Sect. \ref{sec:approach}, we plan to conduct the user study to figure out the exact number of counterfactual samples needed to evaluate models. 
Furthermore, a visualization of all counterfactual generated by current user, nominated with their softmax score, would be helpful to get a better understanding of the model's local behavior.
Finally, we plan to collect and release a large-scale human plausible counterfactual dataset using the crowd-sourcing deployment of our tool.
\vspace{-0.1cm}
\section{Acknowledgment}
This work is partially funded by project MIRROR under grant agreement No. 832921 (project MIRROR from the European Commission: Migration-Related Risks caused by misconceptions of Opportunities and Requirement) and project ROXANNE, the European Union’s Horizon 2020 research and innovation program under grant agreement No. 833635.
\balance
\bibliographystyle{ACM-Reference-Format}
\bibliography{references}

\end{document}